\documentclass{article}

\usepackage{microtype}
\usepackage{graphicx}
\usepackage{subfigure}
\usepackage{booktabs} 

\usepackage{hyperref}

\usepackage[accepted]{icml2023}

\usepackage{amsmath}
\usepackage{amssymb}
\usepackage{mathtools}
\usepackage{amsthm}

\usepackage{xurl}
\usepackage{booktabs}
\usepackage{amsmath}
\usepackage{amsthm}
\usepackage{dsfont}
\usepackage{bbm}

\usepackage{wrapfig}
\usepackage{caption}
\usepackage{multirow}
\usepackage{graphicx}
\usepackage{epsfig}

\usepackage{mathtools}

\newcommand{\vectorproj}[2][]{\textit{proj}_{#1}#2}

\usepackage[capitalize,noabbrev]{cleveref}

\theoremstyle{plain}
\newtheorem{theorem}{Theorem}[section]
\newtheorem{proposition}[theorem]{Proposition}

\theoremstyle{definition}

\theoremstyle{remark}

\usepackage[textsize=tiny]{todonotes}

\icmltitlerunning{\gy{Attributing Image Generative Models using Latent Fingerprints}}

\newif\ifsubmit
\submittrue

\ifsubmit
\newcommand{\ck}[1]{\textcolor{black}{#1}}
\newcommand{\gy}[1]{\textcolor{black}{#1}}
\else
\newcommand{\ck}[1]{{\textcolor{blue}{[ck: #1]}}}
\newcommand{\gy}[1]{\textcolor{blue}{gy: #1}}
\fi

\begin{document}

\twocolumn[
\icmltitle{\gy{Attributing Image Generative Models using Latent Fingerprints}}



\icmlsetsymbol{equal}{*}

\begin{icmlauthorlist}
\icmlauthor{Guangyu Nie}{equal,SEMTE}
\icmlauthor{Changhoon Kim}{equal,SCAI}
\icmlauthor{Yezhou Yang}{SCAI}
\icmlauthor{Yi Ren}{SEMTE}
\end{icmlauthorlist}

\icmlaffiliation{SEMTE}{
 School for Engineering of Matter, Transport and Energy, Arizona State University}
\icmlaffiliation{SCAI}{School of Computing and Augmented Intelligence, Arizona State University}

\icmlcorrespondingauthor{Yezhou  Yang}{yz.yang@asu.edu}
\icmlcorrespondingauthor{Yi Ren }{yren32@asu.edu}

\icmlkeywords{Machine Learning, ICML}

\vskip 0.3in
]
\printAffiliationsAndNotice{\icmlEqualContribution}


\begin{abstract}
Generative models have enabled the creation of contents that are indistinguishable from those taken from nature. 
Open-source development of such models raised concerns about the risks of their misuse for malicious purposes.
One potential risk mitigation strategy is to attribute generative models via fingerprinting. 
Current fingerprinting methods exhibit a significant tradeoff between robust attribution accuracy and generation quality while lacking design principles to improve this tradeoff. 
This paper investigates the use of latent semantic dimensions as fingerprints, from where we can analyze the effects of design variables, including the choice of fingerprinting dimensions, strength, and capacity, on the accuracy-quality tradeoff.
Compared with previous SOTA, our method requires minimum computation and is more applicable to large-scale models. We use StyleGAN2 and the latent diffusion model to demonstrate the efficacy of our method. Codes are available in \href{https://github.com/GuangyuNie/watermarking-through-style-space-edition}{github}.
\end{abstract}

\begin{figure*}[t]
\vspace{-0.1in}
\centering
\includegraphics[width=15cm]{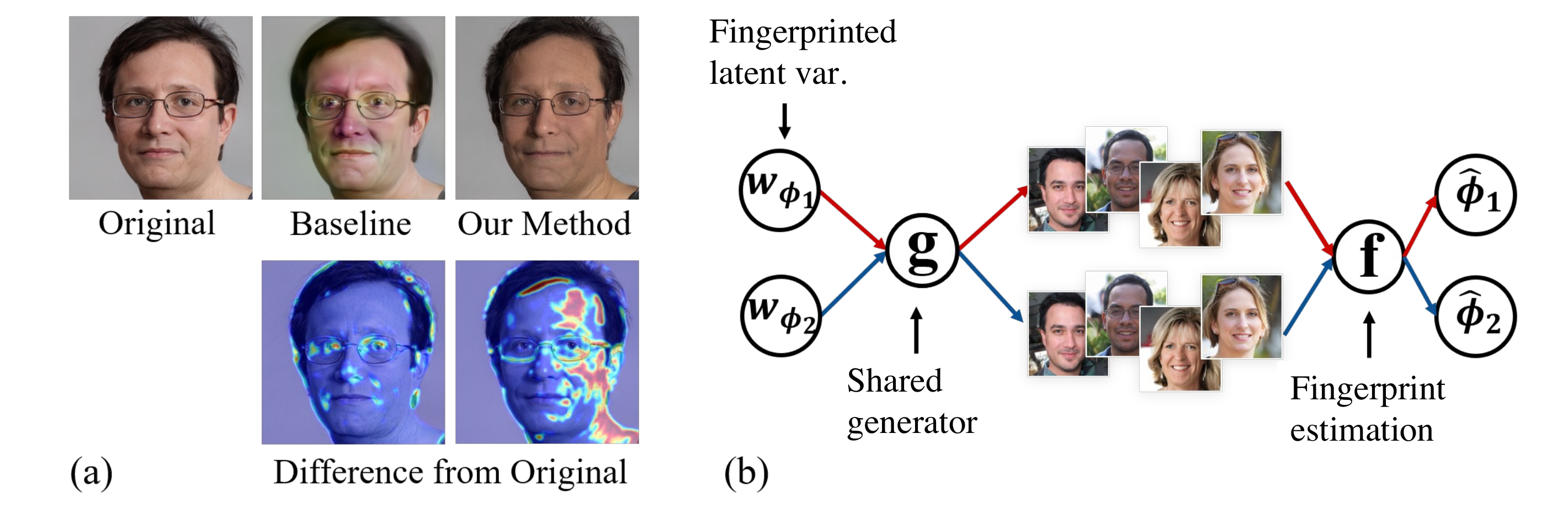}
\vspace{-0.1in}
\caption{
(a) Visual comparison between latent fingerprinting (our method) and shallow fingerprinting~\citep{kim2020decentralized} (baseline). Our method uses subtle semantic changes, rather than strong noises, to maintain attribution accuracy against image postprocesses. 
(b) Schematic of latent fingerprinting: The same generator $g$ and fingerprint estimator $f$ are used for all fingerprinting models. Our method thus requires minimal compute and is scalable to large latent diffusion models.   
}
\vspace{-0.15in}
\label{fig:illustration}
\end{figure*}

\vspace{-0.15in}
\section{Introduction}
\label{submission}
Generative models can now create synthetic content such as images and audio that are indistinguishable from those captured in nature~\citep{karras2020analyzing, Rombach_2022_CVPR, ramesh2022hierarchical, hawthorne2022multi}. 
This poses a serious threat when used for malicious purposes, such as disinformation~\citep{breland_2019} and malicious impersonation~\citep{Satter_2019}. 
Such potential threats have slowed down the industrialization process of generative models, as conservative model inventors hesitate to release their source code~\citep{yu2020responsible}. For example, in 2020, OpenAI refused to release the source code of their GPT-2~\citep{radford2019language} model due to concerns over potential malicious use~\citep{GPT2API}. Additionally, the source code for DALL-E~\citep{ramesh2021zero} and DALL-E 2~\citep{ramesh2022hierarchical} has not been released for the same reason~\citep{Dalle2Risk}.

One potential solution is model attribution~\citep{yu2018attributing, kim2020decentralized, yu2020responsible}, where the model distributor tweaks each user-end model to generate content with model-specific fingerprints. In practice, we consider a scenario where the model distributor or regulator maintains a database of user-specific keys corresponding to each user's downloaded model. In the event of a malicious attempt, the regulator can identify the user responsible for the attempt by using attribution.


Formally, let a set of $n$ generative models be $\mathcal{G}:=\{g_i(\cdot)\}_{i=1}^n$ where $g_i(\cdot): \mathbb{R}^{d_z} \rightarrow \mathbb{R}^{d_x}$ is a mapping from an easy-to-sample distribution $p_z$ to a fingerprinted data distribution $p_{x,i}$ in the content space, and is parameterized by a binary-coded key $\phi_i \in \Phi := \{0,1\}^{d_{\phi}}$. Let $f(\cdot): \mathbb{R}^{d_x} \rightarrow \Phi$ be a mapping that attributes contents to their source models \ck{( Fig.~\ref{fig:illustration}(b))}. We consider four performance metrics of a fingerprinting mechanism: The \textit{attribution accuracy} of $g_i$ is defined as
\begin{equation} \label{eq:acc}
    A(g_i) = \mathbb{E}_{z \sim p_z}\left[ \mathbbm{1}\left(f(g_i(z)) == \phi_i\right) \right].
\end{equation}
The \textit{generation quality} of $g_i$ measures the difference between $p_{x,i}$ and the data distribution used for learning $\mathcal{G}$, e.g., the Fréchet Inception Distance (FID) score~\citep{heusel2017gans} for images. Inception score (IS)~\cite{salimans2016improved} is also measured for $p_{x,i}$ as additional \textit{generation quality} metrics. \textit{Fingerprint secrecy} is measured by the mean structural similarity index measure (SSIM) of individual images drawn from $p_{x,i}$. Compared with generation quality, this metric focuses on how obvious fingerprints are rather than how well two content distributions match. Lastly, the \textit{fingerprint capacity} is $n = 2^{d_{\phi}}$.

Existing model attribution methods exhibit a significant tradeoff between attribution accuracy, generation quality, and secrecy, particularly when countermeasures against deattribution attempts, e.g., image postprocesses, are considered. For example, \citet{kim2020decentralized} use shallow fingerprints for image generators in the form of $g_i(z) = g_0(z) + \phi_i$ where $g_0(\cdot)$ is an unfingerprinted model and show that $\phi_i$s have to significantly alter the original contents to achieve good attribution accuracy against image blurring, causing an unfavorable drop in generation quality and secrecy (Fig.~\ref{fig:illustration}(a)).

To improve this tradeoff, we investigate in this paper latent fingerprints in the form of
$g_i(z) = g_0(\psi(z)+\phi_i)$, where $w := \psi(z) \in \mathbb{R}^{d_w}$ 
contains disentangled semantic dimensions that allow a smoother mapping to the content space~(Fig.~\ref{fig:illustration}(a)). Such $\psi$ has been incorporated in popular models such as StyleGAN (SG)~\citep{karras2019style, karras2020analyzing}, where $w$ is the style vector, and latent diffusion models(LDM)~\citep{Rombach_2022_CVPR}, where $w$ comes from a diffusion process. Existing studies on semantic editing showed that $\mathbb{R}^{d_w}$ consists of linear semantic dimensions~\citep{harkonen2020ganspace,zhu2021low}. Inspired by this, we hypothesize that using subtle yet semantic changes as fingerprints will improve the robustness of attribution accuracy against image postprocesses, and thus investigate the performance of fingerprints that are generated by perturbations along latent dimensions of $\mathbb{R}^{d_w}$. Specifically, we consider latent dimensions as eigenvectors of the covariance matrix of the latent distribution $p_w$, denoted by $\Sigma_w$.

\textbf{Contributions.} (1) \gy{We propose a novel fingerprinting strategy that directly embeds the fingerprints into pretrained generative model, as a mean to achieve responsible white-box model distribution.} (2) We prove and empirically verify that there exists an intrinsic tradeoff between attribution accuracy and generation quality. This tradeoff is affected by fingerprint variables including the choice of the fingerprinting space, the fingerprint strength, and its capacity. Parametric studies on these variables for StyleGAN2 (SG2) and a Latent Diffusion Model(LDM) lead to improved accuracy-quality tradeoff from the previous SOTA. In addition, our method requires negligible computation compared with previous SOTA, rendering it more applicable to popular large-scale models, including latent diffusion ones.
(3) We show that using a postprocess-specific LPIPS metric for model attribution leads to further improved attribution accuracy against image postprocesses.

\vspace{-0.05in}
\section{Related Work}
\vspace{-0.05in}
\textbf{Model attribution through fingerprint encoding and decoding.} \citet{yu2020responsible} propose to encode binary-coded keys into images through $g_i(z) = g_0([z, \phi_i])$ and to decode them via another learnable function. This requires joint training of the encoder and decoder over $\mathbb{R}^{d_z} \times \Phi$ to empirically balance attribution accuracy and generation quality. 
Since fingerprint capacity is usually high (i.e., $2^{d_{\phi}}$), training is made tractable by sampling only a small subset of fingerprints. Thus this method is computationally expensive and lacks a principled understanding of how the fingerprinting mechanism affects the accuracy-quality tradeoff. In contrast, our method does not require any additional training and mainly relies on simple principle component analysis of the latent distribution.

\textbf{Certifiable model attribution through shallow fingerprints.} \citet{kim2020decentralized} propose shallow fingerprints $g_i(z) = g_0(z) + \phi_i$ and linear classifiers for attribution. These simplifications allow the derivation of sufficient conditions of $\Phi$ to achieve certifiable attribution of $\mathcal{G}$. Since the fingerprints are added as noises rather than semantic changes coherent with the generated contents, 
increased noise strength 
becomes necessary to maintain attribution accuracy under postprocesses. While this paper does not provide attribution certification for latent fingerprint, we discuss the technical feasibility and challenges in achieving this goal. 

\textbf{StyleGAN and low-rank subspace.} Our study focuses on popular image generation models which share an architecture rooted in SG: A Gaussian distribution is first transformed into a latent distribution ($p_w$), samples from which are then decoded into images. 
\citet{harkonen2020ganspace} apply principal component analysis on $p_w$ distribution and found semantically meaningful editing directions.
\citet{zhu2021low} use local Jacobian ($\nabla_w g$) to derive perturbations that enable local semantic editing of generated images, and show that such semantic dimensions are shared across the latent space. 
In this study, we show that the mean of the Gram matrix for local editing ($\mathbb{E}_{w \sim p_w}[\nabla_w g^T \nabla_w g]$) and the covariance of $w$ ($\Sigma_w$) are qualitatively similar in that both reveal major to minor semantic dimensions through their eigenvectors. 

\textbf{GAN inversion.} 
The model attribution problem can be formulated as a GAN inversion problem. 
A learning-based inversion~\citep{perarnau2016invertible, bau2019inverting} optimizes parameters of the encoder network which map an image to latent code $z$.
On the other hand, optimization-based inversion~\citep{abdal2019image2stylegan, huh2020transforming} solve for latent code $z$ that minimizes distance metric between a given image and generated image $g(z)$.
The learning-based method is computationally more efficient in the inference stage compared to optimization-based method. However, optimization-based GAN inversion achieves a superior quality of latent interpretation, which can be referred to as the quality-time tradeoff~\citep{xia2022gan}.
In our method, we utilized the optimization-based inversion, as faithful latent interpretation is critical in our application. To further enforce faithful latent interpretation, 
we incorporate existing techniques, e.g., parallel search, to solve this non-convex problem, but uniquely exploit the fact that fingerprints are small latent perturbations to enable analysis on the accuracy-quality tradeoff.


    

\begin{table*}[h]
  \caption{
  Attribution accuracy and generation quality of the proposed method.  FID-$g_0$ and IS-$g_0$ represent the baseline FID and inception score for the image quality of generative models prior to fingerprinting. $\uparrow$ ($\downarrow$) indicates higher (lower) is desired. 
  The standard deviation is in parentheses.
  }
  \label{Tab:Accuracy}
  \centering
  \begin{tabular}{cccccccccc}
    \toprule
    Model & Dataset & \multicolumn{3}{c}{Attribution Accuracy} & \multicolumn{5}{c}{Image Quality}\\
    \cmidrule(lr){3-5}    \cmidrule(lr){6-10}
     &  & Ours & w/o $\alpha$-reg & w/o LHS  & FID-$g_0$ & FID$\downarrow$&IS-$g_0$ $\uparrow$ &IS   $\uparrow$ &SSIM $\uparrow$\\
    
    \midrule
    SG2 & FFHQ & \textbf{0.983} & 0.877 & 0.711 & 7.24 & 8.59 & 5.16&4.96
    & 0.93(0.02) \\
    SG2 & AFHQ Cat & \textbf{0.993} & 0.991 & 0.972  & 6.35& 7.87&1.65&2.32 &0.97(0.01)\\
    SG2 & AFHQ Dog & \textbf{0.999} & 0.998 & 0.981 & 3.80& 5.36& 9.76&12.33&0.95(0.01)\\
    LDM & FFHQ & \textbf{0.996} & 0.364 & 0.872 & 12.34 & 13.63 & 4.50&4.35 &0.94(0.01)\\
    \bottomrule
  \end{tabular}
\end{table*}

\vspace{-0.05in}
\section{Methods}\label{sec:methods}

\subsection{Notations and preliminaries} 
\vspace{-0.05in}


\textbf{Notations.} For $x \in \mathbb{R}^n$ and $A \in \mathbb{R}^{n \times m}$, denote by $\vectorproj[A]{x}$ the projection of $x$ to $span(A)$, and by $A^{\dagger}$ the pseudo inverse of $A$. For parameter $a$, we denote by $\hat{a}$ its estimate and $\epsilon_a = \hat{a} - a$ the error. $\nabla_x f$ is the gradient of $f$ with respect to $x$, $\mathbb{E}_{x \sim p_x}[\cdot]$ is an expectation over $p_x$, and $tr(B)$ (resp. $det(B)$) is the trace (resp. determinant) of $B \in \mathbb{R}^{n \times n}$. $diag(\lambda) \in \mathbb{R}^{n \times n}$ diagonalizes $\lambda \in \mathbb{R}^{n}$.

\textbf{Latent Fingerprints.} Contemporary generative models, e.g., SG2 and LDM, consist of a disentanglement mapping $\psi: \mathcal{R}^{d_z} \rightarrow \mathcal{R}^{d_w}$ from an easy-to-sample distribution $p_z$ to a latent distribution $p_w$, followed by a generator $g: \mathcal{R}^{d_w} \rightarrow \mathcal{R}^{d_x}$ that maps $w$ to the content space. 
In particular, $\psi$ is a multilayer perception network in SG2 and a diffusion process in a diffusion model. 
Existing studies showed that linear perturbations along principal components of $\nabla_w g$ enable semantic editing, and such perturbation directions are often applicable over $w \sim p_w$~\citep{harkonen2020ganspace, zhu2021low}. Indeed, instead of local analysis on the Jacobian, \cite{harkonen2020ganspace} showed that principal component analysis directly on $p_w$ also reveals semantic dimensions. This paper follows these existing findings and uses a subset of semantic dimensions as fingerprints. Specifically,
let $U \in \mathbb{R}^{d_w \times (d_w-d_{\phi})}$ and $V \in \mathbb{R}^{d_w \times d_{\phi}}$ be orthonormal and complementary. Given random seed $z \in \mathbb{R}^{d_z}$, user-specific key $\phi \in \mathbb{R}^{d_{\phi}}$, and strength $\sigma \in \mathbb{R}$, let $\alpha =  U^{\dagger}\vectorproj[U]{\psi(z)} \in \mathbb{R}^{d_w-d_{\phi}}$, the fingerprinted latent variable is
\begin{equation}
    w_{\phi}(\alpha) = U\alpha + \sigma V \phi,
\end{equation}
where $\alpha \sim p_{\alpha}$ and $p_{\alpha}$ is induced by $p_w$. 
Then, the user can generate fingerprinted images, $g(w_\phi(\alpha))$.
The choice of $(U, V)$ and $\sigma$ affects the attribution accuracy and generalization performance, which we analyze in Sec.~\ref{sec:acc-quality-tradeoff}.

\textbf{Attribution.}To decode user-specific key from the image $g(w_\phi(\alpha))$, we formulate an optimization problem:
\begin{equation*}
\begin{aligned}
      \min_{\hat{\alpha}, \hat{\phi}} \quad &  l\left(g(w_{\hat{\phi}}(\hat{\alpha})), g(w_{\phi}(\alpha))\right) \\
      s.t. \quad & \hat{\alpha}_i \in [\alpha_{l,i}, \alpha_{u,i}], \quad \forall i=1,...,d_{w} - d_{\phi}.
\end{aligned}
\end{equation*}
\ck{While $l$ is $l_2$ norm for analysis in Sec.\ref{sec:acc-quality-tradeoff}, here we minimize LPIPS~\citep{zhang2018unreasonable} which measures the perceptual difference between two images.}
Through experiments, we discovered that attribution accuracy can be improved by constraining $\alpha$.
Here the upper and lower bounds of $\alpha$ are chosen based on the empirical limits observed from $p_{\alpha}$. 
In practice, we introduce a penalty on $\hat{\alpha}$ with large enough Lagrange multipliers and solve the resulting unconstrained problem. To avoid convergence to unfavorable local solutions, we also employ parallel search with n initial guesses of $\hat{\alpha}$ drawn through Latin hypercube sampling (LHS). 

\subsection{Accuracy-quality tradeoff}\label{sec:acc-quality-tradeoff}
\vspace{-0.05in}

\textbf{Attribution accuracy.} 
Define $J_{w} = \nabla g(w)$, 
$H_{w} = J_{w}^TJ_{w}$. Let $\bar{H}_{w} = \mathbb{E}_{w \sim p_w} [H_{w}]$ be the mean Gram matrix, and $\bar{H}_{\phi} = \mathbb{E}_{\alpha \sim p_\alpha} [H_{w_{\phi}(\alpha)}]$ be its fingerprinted version. Let $l: \mathbb{R}^{d_x} \times \mathbb{R}^{d_x} \rightarrow \mathbb{R}$ be a distance metric between two images, $(\hat{\alpha}, \hat{\phi})$ the estimates.
To analyze how $(V, U)$ affects the attribution accuracy, we use the following simplifications and assumptions: 
\textbf{(A1)} $l(\cdot,\cdot)$ is the $l_2$ norm.
\textbf{(A2)} Both $||\epsilon_{\alpha}||$ and $\sigma$ are small. In practice we achieve small $||\epsilon_{\alpha}||$ through parallel search (see Appendix~\ref{app:error_alpha}). \textbf{(A3)} Since our focus is on $\epsilon_{\phi}$, we further assume that the estimation of $\alpha$, denoted by $\hat{\alpha}(\alpha)$, is independent from $\phi$, and $\epsilon_{\alpha}$ is constant. This allows us to ignore the subroutine for computing $\hat{\alpha}(\alpha)$ and turns the estimation problem into an optimization with respect to only $\epsilon_{\phi}$. 
Formally, we have the following proposition (see Appendix~\ref{appendix:proposition1} for proof):
\begin{proposition}\label{proposition_1}
$\exists ~c>0$ such that if $\sigma \leq c$ and $||\epsilon_{\alpha}||_2 \leq c$, the fingerprint estimation problem
\begin{equation*} \label{eq:obj}
    \min_{\hat{\phi}} \quad \mathbb{E}_{\alpha \sim p_{\alpha}} \left[\|g(w_{\hat{\phi}}(\hat{\alpha}(\alpha))) - g(w_{\phi}(\alpha))\|_2^2\right]
\end{equation*}
has an error $\epsilon_{\phi} = -(\sigma^2 V^T \bar{H}_{\phi} V)^{-1}V^T \bar{H}_{\phi} U \epsilon_{\alpha}$.
    
\end{proposition}

\textbf{Remarks:} (1) Similar to the classic design of the experiment, one can reduce $||\epsilon_{\phi}||$ by maximizing $det(V^T \bar{H}_{\phi} V)$, which sets columns of $V$ as the eigenvectors associated with the largest $d_{\phi}$ eigenvalues of $\bar{H}_{\phi}$. However, $\bar{H}_{\phi}$ is neither computable because $\phi$ is unknown during the estimation, nor is it tractable because $J_{w_{\phi}(\alpha)}$ is large in practice. To this end, we propose to use the covariance of $p_w$, denoted by $\Sigma_w$, to replace $\bar{H}_{\phi}$ in experiments. In Appendix~\ref{app:similar}, we support this approximation empirically by showing that $\Sigma_w$ and $\bar{H}_{w}$ (the non-fingerprinted mean Gram matrix) are qualitatively similar in that the principal components of both matrices offer disentangled semantic dimensions.
(2) Let the $k$th largest eigenvalue of $\bar{H}_{w}$ be $\gamma_k$. By setting columns of $V$ as the eigenvectors of $\bar{H}_{w}$ associated with the largest $d_{\phi}$ eigenvalues, and by noting that $\hat{\phi}$ is accurate only when all of its elements match with $\phi$ (\eqref{eq:acc}), the worst-case estimation error is governed by $\gamma_{d_{\phi}}^{-1}$. This means that higher key capacity, i.e., larger $d_{\phi}$, leads to worse attribution accuracy. 
(3) From the proposition, $\epsilon_{\phi} = 0$ if $V$ and $U$ are complementary sets of eigenvectors of $\bar{H}_{\phi}$. In practice this decoupling between $\epsilon_{\phi}$ and $\epsilon_{\alpha}$ cannot be achieved due to the assumptions and approximations we made. 




\textbf{Generation quality.} For analysis purpose, we approximate the original latent distribution $p_w$ by $w = \mu + U\alpha + V\beta$ where $\alpha \sim \mathcal{N}(0,diag(\lambda_U))$, $\beta \sim \mathcal{N}(0,diag(\lambda_V))$, and $\mu = \mathbb{E}_{w \sim p_w}[w]$. $\lambda_U \in \mathbb{R}^{d_w - d_{\phi}}$ and $\lambda_V \in \mathbb{R}^{d_{\phi}}$ are calibrated to match $p_w$. Denote $\lambda_{V,max} = \max_i \{\lambda_{V,i}\}$. 
A latent distribution fingerprinted by $\phi$ is similarly approximated as $w_{\phi} = \mu + U\alpha + \sigma V\phi$. With mild abuse of notation, let $g$ be the mapping from the latent space to a feature space (usually defined by an Inception network in FID) and is continuously differentiable. Let the mean and covariance matrix of $w_i$ be $\mu_i$ and $\Sigma_i$, respectively. 
Denote by $\bar{H}_U = \mathbb{E}_{\alpha}[J_{\mu + U\alpha}^T J_{\mu + U\alpha}]$ the mean Gram matrix in the subspace of $U$, and let $\gamma_{U, max}$ be the largest eigenvalue of $\bar{H}_U$. 
We have the following proposition to upper bound $||\mu_0 - \mu_1||_2^2$ and $|tr(\Sigma_0) - tr(\Sigma_1)|$, both of which are related to the FID score for measuring the generation quality (see Appendix~\ref{appendix:proposition2} for proof):

\begin{proposition}\label{proposition_2}
For any $\tau >0$ and $\eta \in (0,1)$, $\exists ~c(\tau, \eta)>0$ and $\nu>0$, such that if $\sigma \leq c(\tau, \eta)$ and $\lambda_{V,i} \leq c(\tau, \eta)$ for all $i=1,...,d_{\phi}$, then $\|\mu_0 - \mu_1\|_2^2 \leq \sigma^2\gamma_{U, max} d_{\phi} + \tau$ and $|tr(\Sigma_0 - \Sigma_1)| \leq \lambda_{V, max} \gamma_{U, max} d_{\phi} + 2\nu \sigma \sqrt{d_{\phi}} + \tau$ with probability at least $1-\eta$.
\end{proposition}

\textbf{Remarks:} Recall that for improving attribution accuracy, a practical approach is to choose $V$ as eigenvectors associated with the largest eigenvalues of $\Sigma_w$. Notices that with the approximated distribution with $\alpha \sim \mathcal{N}(0,diag(\lambda_U))$ and $\beta \sim \mathcal{N}(0,diag(\lambda_V))$, $\Sigma_w = diag([\lambda_U^T, \lambda_V^T]^T)$. On the other hand, from Proposition~\ref{proposition_2}, generation quality improves if we minimize $\lambda_{V,max}$ by choosing $V$ according to the smallest eigenvalues of $\Sigma_w$. In addition, smaller key capacity ($d_{\phi}$) and lower strength ($\sigma$) also improve the generation quality. Propositions~\ref{proposition_1} and~\ref{proposition_2} together reveal the intrinsic accuracy-quality tradeoff. 

\begin{figure*}%
\vspace{-0.05in}
    \centering
    \includegraphics[width=15cm]{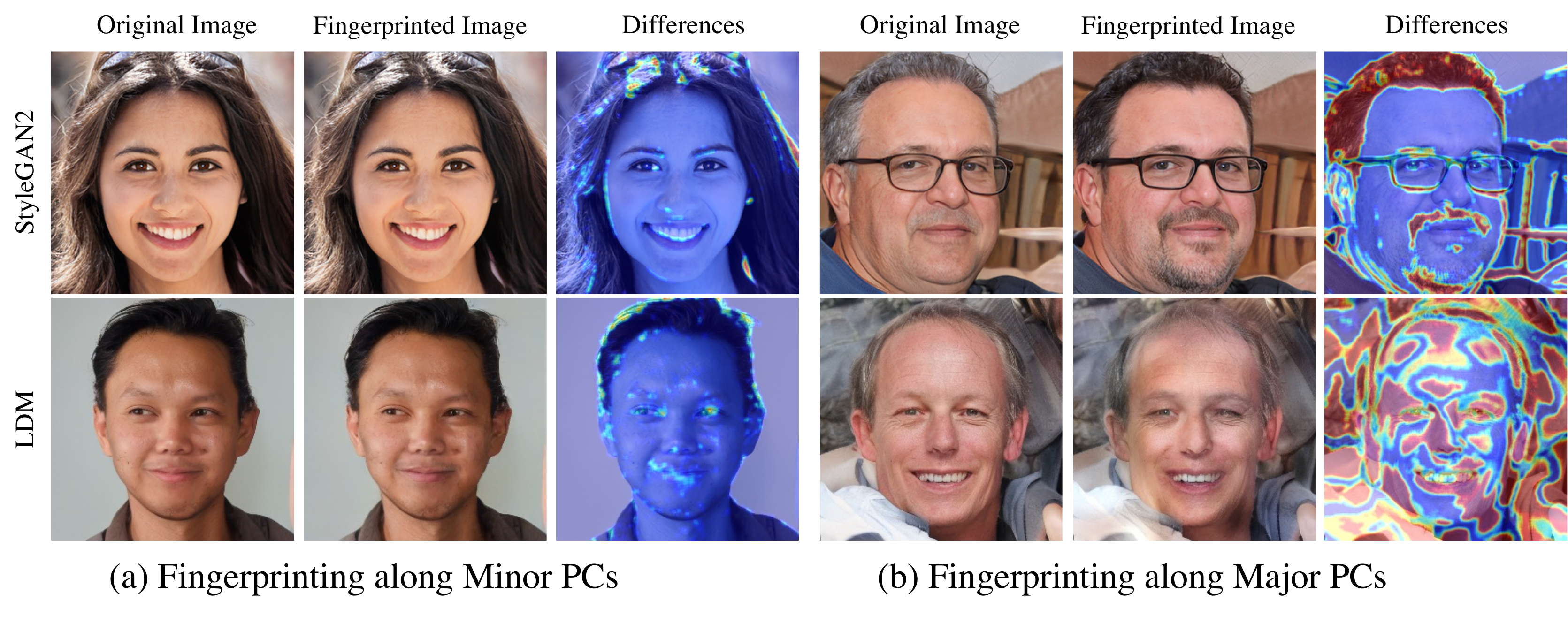}
    \vspace{-0.1in}
    \caption{Visualization of fingerprints along minor and major principal components of the covariance of the latent distribution. (Top) StyleGAN2. (Bottom) Latent Diffusion Model (LDM).}%
    \label{fig:example_of_SG2}%
    \vspace{-0.15in}
\end{figure*}

\vspace{-0.05in}
\section{Experiments} \label{sec:experiment}
\vspace{-0.05in}
In this section, we present empirical evidence of the accuracy-quality tradeoff and show an improved tradeoff from the previous SOTA by using latent fingerprints. Experiments are conducted for both with and without a combination of postprocesses including (image noising, blurring, and JPEG compression, and their combination).  

\subsection{Experiment settings}
\vspace{-0.05in}

\textbf{Models, data, and metrics.} We conduct experiments on SG2~\citep{karras2020analyzing} and LDM~\citep{Rombach_2022_CVPR} models trained on various datasets including FFHQ~\citep{karras2019style}, AFHQ-Cat, and AFHQ-Dog~\citep{choi2020stargan}. 
Generation quality is measured by the Frechet Inception distance (FID)~\citep{heusel2017gans} and inception score (IS)~\citep{salimans2016improved}, attribution accuracy by \eqref{eq:acc}, and fingerprint secrecy by SSIM.

\textbf{Latent fingerprint dimensions.} 
\ck{To approximate $\Sigma_w$, we drew 10K samples from $p_w$ for SG2, which has a semantic latent space dimensionality of 512, and 50K samples from $p_w$ for LDM, which has a semantic latent space dimensionality of 12,288.}
We define fingerprint dimensions $V$ as a subset eigenvectors of $\Sigma_w$ associated with consecutive eigenvalues: $V := PC[i:j]$, where $PC$ is the full set of principal components of $\Sigma_w$ sorted by their variances in the descending order while $i$ and $j$ represent the starting and ending indices of the subset.


\textbf{Attribution.}
To compute the empirical accuracy~(Eq.~\eqref{eq:acc}), we use 1K samples drawn from $p_z$ for each fingerprint $\phi$, and use 1K fingerprints where each bit is drawn independently from a Bernoulli distribution with $p=0.5$.  
In Table~\ref{Tab:Accuracy}, we show that both constraints on $\hat{\alpha}$ and parallel search with 20 initial guesses improve the empirical attribution accuracy across models and datasets. Notably, constrained estimation is essential for the successful attribution of LDMs. In these experiments, $V$ is chosen as the eigenvectors associated with the 64 smallest eigenvalues of $\Sigma_w$ as a worst-case scenario for attribution accuracy.

\subsection{Attribution performance without postprocessing}
\vspace{-0.05in}
We present generation quality results in Table~\ref{Tab:Accuracy}. Since the least variant principal components are used as fingerprints, generation quality (FID) and fingerprint secrecy (SSIM) are preserved. 

The results suggest that the attribution accuracy, generation quality, capacity ($2^{64}$), and fingerprint secrecy are all acceptable using the proposed method.
Fig.~\ref{fig:example_of_SG2} visualizes and compares latent fingerprints generated from small vs. large eigenvalues of $\Sigma_w$. Fingerprints corresponding to small eigenvalues are non-semantic, while those to large eigenvalues create semantic changes. We will later show that semantic yet subtle (perceptually insignificant) fingerprints are necessary to counter image postprocesses.

\textbf{Accuracy-quality tradeoff.} 
Table~\ref{tab:trade-off} summarizes the tradeoff when we vary the choice of $V$ and the fingerprint strength $\sigma$ while fixing the fingerprint length $d_{\phi}$ to 64. Then in Table~\ref{Tab:key_length} we sweep $d_{\phi}$ while keeping $V$ as PCs associated with smallest eigenvalues of $\Sigma_w$ and $\sigma=1$. The experiments are conducted on SG2 and LDM on the FFHQ dataset.
The empirical results in Table~\ref{tab:trade-off} are consistent with our analysis: Accuracy decreases while generation quality improves when $V$ is moved from major to minor principal components. 
For fingerprint strength, however, we observe that the positive effect of strength on the accuracy, as predicted by Proposition~\ref{proposition_1}, is only limited to small $\sigma$. This is because larger $\sigma$ causes pixel values to go out of bounds, causing loss of information.
In Table~\ref{Tab:key_length}, we summarize the attribution accuracy, FID, and SSIM score under 32- to 128-bit keys. Accuracy and generation quality, in particular the latter, are both affected by $d_{\phi}$ as predicted. 

\begin{table*} [t]
\vspace{-0.05in}
  \caption{
  Tradeoff between attribution accuracy (Att.) and generation quality (FID, IS) under different fingerprinting directions (PC) and strength ($\sigma$).
  }
  \label{tab:trade-off}
  \centering
  \renewcommand{\tabcolsep}{5pt}
  \begin{tabular}{llllllllllll}
    \toprule
    StyleGAN2
     &\multicolumn{3}{c}{$\sigma = 0.6$} &  \multicolumn{3}{c}{$\sigma = 1.0$} & \multicolumn{3}{c}{$\sigma=6.0$} & \\
     \cmidrule(lr){2-4}    \cmidrule(lr){5-7} \cmidrule(lr){8-10}
      & Att.~$\uparrow$ & FID~$\downarrow$&IS $\uparrow$ & Att.~$\uparrow$ & FID~$\downarrow$&IS  $\uparrow$ & Att.~$\uparrow$& FID~$\downarrow$&IS  $\uparrow$\\ 
    \midrule
    PC[0:64]  & 0.99 & 129.0 &1.23 &0.99 & 110.8 &1.59& 0.99 & 101.3&4.31\\
    PC[128:192]  & 0.98 & 8.5&4.93  &0.99 & 8.7& 4.92 & 0.99 & 39.2&3.94\\
    PC[256:320] & 0.98 & 8.6&4.96 &0.99 & 9.1 &4.87& 0.96 & 31.1&3.90\\
    PC[448:512]  & 0.97 & 8.1&4.99  &0.98 & 8.5&4.96 & 0.90 & 26.3 &4.75\\
    \midrule
    \midrule
    LDM
    &\multicolumn{3}{c}{$\sigma = 1.0$} &  \multicolumn{3}{c}{$\sigma = 2.0$} & \multicolumn{3}{c}{$\sigma=3.0$} & \\
         \cmidrule(lr){2-4}    \cmidrule(lr){5-7} \cmidrule(lr){8-10}
      & Att.~$\uparrow$ & FID~$\downarrow$&IS $\uparrow$ & Att.~$\uparrow$ & FID~$\downarrow$ &IS $\uparrow$& Att.~$\uparrow$& FID~$\downarrow$&IS  $\uparrow$\\ 
    \midrule
    PC[0:64]  &   0.99 & 33.62 &3.84  &0.99 & 33.17 &3.70 & 0.99 & 34.07 &3.82\\
    PC[1000:1064]  & 0.77 & 13.32  &4.37 &0.99 & 13.75 &4.40 & 0.99 & 16.03 &4.35\\
    PC[2000:2064] & 0.32 & 13.17 &4.45 &0.99 & 13.63 &4.35 & 0.99 & 15.74 &4.34\\
    PC[3000:3064] & 0.12 & 12.98 &4.43 &0.97 & 13.61 &4.45 & 0.99 & 15.44 &4.48\\
    PC[4000:4064]  &0.00  & 12.77 &4.35  &0.96 & 13.61  &4.42 & 0.99 & 15.41 &4.41\\
    \bottomrule
  \end{tabular}
  \vspace{-0.1in}
\end{table*}

\begin{table}[h]
  \caption{
  Attribution accuracy (Att.) and generation quality for different key lengths ($d_{\phi}$). FID-BL is the baseline FID score. $\uparrow$ ($\downarrow$) indicates higher (lower) is desired. The standard deviation of SSIM is in parentheses.
  }
  \label{Tab:key_length}
  \centering
  \begin{tabular}{lllllll}
    \toprule
    $d_{\phi}$
    & Att. $\uparrow$ &FID-BL & FID $\downarrow$ &SSIM $\uparrow$ &IS  $\uparrow$ \\
    \midrule
    32 &0.982 & 7.24 & 8.49  &0.96(0.01)& 4.90  \\
    64 &0.983 & 7.24 & 8.59  & 0.92(0.02)& 4.96 \\
    96 &0.981 & 7.24 & 9.50  & 0.90(0.02)& 4.93 \\
    128 &0.973 & 7.24 & 9.61  & 0.87(0.03)& 4.91 \\
    \bottomrule
\vspace{-0.20in}
  \end{tabular}
\end{table}

\subsection{Fingerprint performance with postprocessing}
\vspace{-0.05in}

We now consider more realistic scenarios where generated images are postprocessed, either maliciously as an attempt to remove the fingerprints or unintentionally, before they are attributed. Under this setting, our method achieves better accuracy-quality tradeoff than shallow fingerprinting under two realistic settings: (1) when noising and JPEG compression are used as \textit{unknown} postprocesses, and (2) when \textit{the set of postprocesses}, rather than the ones that are actually chosen, is known. 

\textbf{Postprocesses.} To keep our solution realistic, we solve the attribution problem by assuming that the potential postprocesses are unknown:
\begin{equation*} \label{eq:estimation}
\begin{aligned}
      \min_{\hat{\alpha}, \hat{\phi}} \quad &  l\left(g(w_{\hat{\phi}}(\hat{\alpha})), T(g(w_{\phi}(\alpha)))\right) \\
      s.t. \quad & \hat{\alpha}_i \in [\alpha_{l,i}, \alpha_{u,i}], \quad \forall i=1,...,d_{w} - d_{\phi}.  
\end{aligned}
\end{equation*}
where $T: R^{d_x} \rightarrow R^{d_x}$ is a postprocess function, and $T(g(w_{\phi}(\alpha)))$ is a given image from which the fingerprint is to be estimated.
We assume that $T$ does not change the image in a semantically meaningful way, because otherwise the value of the image for either an attacker or a benign user will be lost. Since our method adds semantically meaningful perturbations to the images, we expect such latent fingerprints to be more robust to postprocesses than shallow ones~\cite{kim2020decentralized} added directly to images and will lead to improved attribution accuracy. To test this hypothesis, we consider four types of postprocesses: \texttt{Noising}, \texttt{Blurring}, \texttt{JPEG}, and \texttt{Combo}. 
\texttt{Noising} adds a Gaussian white noise of standard deviation randomly sample from $U$[0, 0.1]. 
\texttt{Blurring} uses a randomly selected Gaussian kernel size from [3, 7, 9, 16, 25] and a standard deviation of [0.5, 1.0, 1.5, 2.0].
We randomly sample the \texttt{JPEG} quality from [80, 70, 60, 50].
These parameters are chosen to be mild so that images do not lose their semantic contents. 
And \texttt{Combo} randomly chooses a subset of the three through a binomial distribution with $p=0.5$ and uses the same postprocess parameters.

\begin{table*} [t]
\vspace{-0.05in}
  \caption{Comparison on accuracy-quality tradeoff between proposed and baseline methods under image postprocesses. \ck{The experiments are tested using StyleGAN2-FFHQ.} Fingerprinting strength $\sigma = 3$. The FID score of the baseline method is 96.24. KN (UK) stands for when attribution accuracy is measured with (without) the knowledge of the attack. The standard deviation is in parentheses.}
  
  \label{tab:robustness}
  \centering
  \renewcommand{\tabcolsep}{5pt}
  \begin{tabular}{llllllllll}
    \toprule
    Metric & Model &\multicolumn{2}{c}{\texttt{Blurring}} &  \multicolumn{2}{c}{\texttt{Noising}} & \multicolumn{2}{c}{\texttt{JPEG}} & \multicolumn{2}{c}{\texttt{Combo}}\\
    -      & -  & UK & KN & UK & KN & UK & KN & UK & KN \\ 
    \midrule
    \multirow{3}{*}{Att. $\uparrow$} 
    & BL~\cite{kim2020decentralized}  & 0.85 & 0.88  &0.85 & 0.87 & 0.87 & 0.89 &0.83 &0.88 \\
    & PC[0:32]  & 0.79 & 0.99  &0.99 & 0.99 & 0.98 & 0.99 &0.82 &0.99 \\
    & PC[16:48] & 0.56 &0.92 &0.95 & 0.99 & 0.98 & 0.99 & 0.42 & 0.88\\
    & PC[32:64]  & 0.32 & 0.83  &0.93 & 0.98 & 0.98 & 0.99 &0.26 &0.79 \\

    \midrule
     \multirow{3}{*}{SSIM $\uparrow$}    
     & BL~\cite{kim2020decentralized}
     &\multicolumn{2}{l}{0.67(0.08)} 
     &\multicolumn{2}{l}{0.68(0.07)}  
     &\multicolumn{2}{l}{0.67(0.07)} 
     &\multicolumn{2}{l}{0.66(0.06)}  \\
     
     & PC[0:32] 
     & \multicolumn{2}{l}{0.27(0.04)} 
     &\multicolumn{2}{l}{0.27(0.04)}  
     &\multicolumn{2}{l}{0.27(0.04)} 
     &\multicolumn{2}{l}{0.27(0.04)}  \\
   
    & PC[16:48]
     &\multicolumn{2}{l}{0.40(0.08)}
     &\multicolumn{2}{l}{0.40(0.08)}
     &\multicolumn{2}{l}{0.40(0.08)}
     &\multicolumn{2}{l}{0.40(0.08)}\\
    
    & PC[32:64]
     &\multicolumn{2}{l}{0.56(0.07)}  
     &\multicolumn{2}{l}{0.56(0.07)} 
     &\multicolumn{2}{l}{0.56(0.07)} 
     &\multicolumn{2}{l}{0.56(0.07)} \\
    
    \midrule
    
     \multirow{3}{*}{IS $\uparrow$}  
    
     & BL~\cite{kim2020decentralized}
     & \multicolumn{2}{l}{2.86} 
     &\multicolumn{2}{l}{3.02}  
     &\multicolumn{2}{l}{2.91}  
     &\multicolumn{2}{l}{2.90}  \\
     
     & PC[0:32] 
     & \multicolumn{2}{l}{2.93} &\multicolumn{2}{l}{2.93}  &\multicolumn{2}{l}{2.93}  &\multicolumn{2}{l}{2.93}  \\
    & PC[16:48] 
    &\multicolumn{2}{l}{4.35}    
     &\multicolumn{2}{l}{4.35}
     &\multicolumn{2}{l}{4.35}
     &\multicolumn{2}{l}{4.35}     \\
     
    & PC[32:64]  &\multicolumn{2}{l}{4.50}  &\multicolumn{2}{l}{4.50}  &\multicolumn{2}{l}{4.50}	 &\multicolumn{2}{l}{4.50} \\

    \midrule
     \multirow{3}{*}{FID $\downarrow$}  
     & BL~\cite{kim2020decentralized}
     & \multicolumn{2}{l}{99.05} 
     &\multicolumn{2}{l}{93.04}  
     &\multicolumn{2}{l}{97.70}  
     &\multicolumn{2}{l}{100.15}  \\
     
     & PC[0:32] & 
     \multicolumn{2}{l}{102.26} &\multicolumn{2}{l}{102.26}  &\multicolumn{2}{l}{102.26}  &\multicolumn{2}{l}{102.26}  \\
    & PC[16:48] 
    &\multicolumn{2}{l}{31.25}    
     &\multicolumn{2}{l}{31.25}
     &\multicolumn{2}{l}{31.25}
     &\multicolumn{2}{l}{31.25}     \\
     
    & PC[32:64]  &\multicolumn{2}{l}{27.50}  &\multicolumn{2}{l}{27.50}  &\multicolumn{2}{l}{27.50}	 &\multicolumn{2}{l}{27.50} \\
    
    \bottomrule
    
  \end{tabular}
  \vspace{-0.1in}
\end{table*}

\begin{table*}[h]
  \caption{
  Accuracy-quality tradeoff under \texttt{Combo} attack. $V$ is defined as the 8, 16, and 32 eigenvectors of $\Sigma_w$ starting from the 33rd eigenvectors. $\sigma=3$. KN (UK) stands for when attribution accuracy is measured with (without) knowledge of attacks. The standard deviation is in parentheses.
  }
  \label{Tab:combination_sweep}
  \centering
  \begin{tabular}{cccccccc}
    \toprule
    Model &  Key length & UK & KN  &SSIM $\uparrow$ &IS $\uparrow$  & FID$\downarrow$  \\
    
    \midrule
    BL~\cite{kim2020decentralized} & N/A & 0.83 & 0.88 &0.66(0.06) &2.90 & 100.15 \\
    PC[32:40] & 8 & 0.65 & \textbf{0.89}  &\textbf{0.73(0.06)} &4.75 & \textbf{12.35}\\
    PC[32:48] & 16 & 0.45 & 0.85 &0.65(0.06)  &\textbf{4.86} & 13.25\\
    PC[32:64] & 32 & 0.26 & 0.79  &0.57(0.07) &4.50 &27.50\\
    \bottomrule
  \end{tabular}
\end{table*}

    

\begin{figure*}
\centering
\includegraphics[width=15cm]{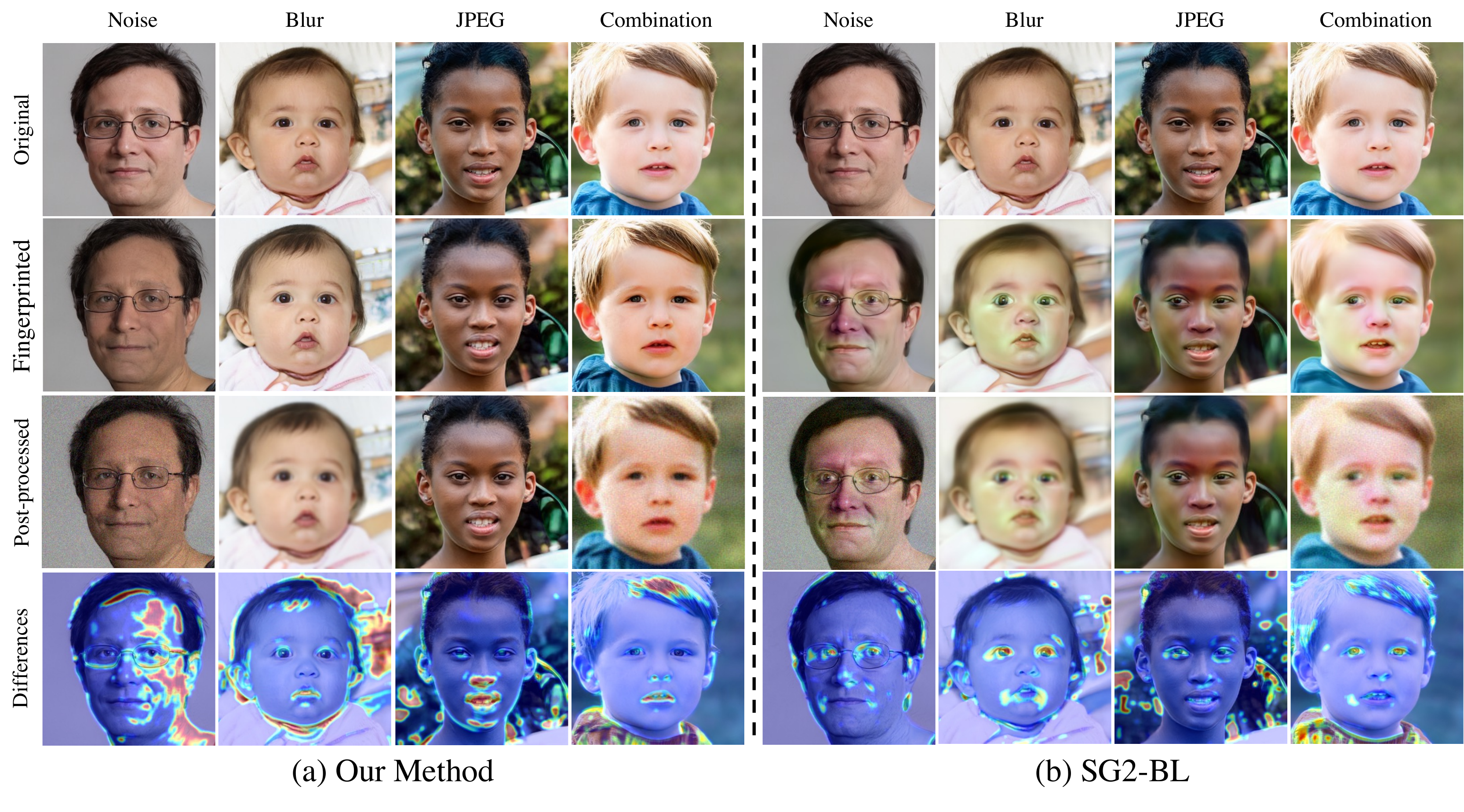}
\vspace{-0.1in}
\caption{\textbf{Comparison on generation quality between our method and the baseline with similar attribution accuracy} The first row shows original images generated without fingerprinting.
Each image in the second row represent robustly fingerprinted images against corresponding post-processes.
The next row illustrates post-processed images.
The last row depicts the differences between the original (the first row) and fingerprinted images (the second row) using a heat map. 
Even if our method shows large pixel value changes, the fingerprints are not perceptible compared with the baseline method (see second row). 
}
\label{fig:robustness}
\vspace{-0.15in}
\end{figure*}

\textbf{Modified LPIPS metric.} 
In addition to testing the worst-case scenario where postprocesses are completely unknown, we also consider cases  where they are known. While this is unrealistic for individual postprocesses, it is worth investigating when we assume that the set of postprocesses, rather than the ones that are actually chosen, is known. Within this scenario, we show that modifying LPIPS according to the postprocess improves the attribution accuracy. To explain, LPIPS is originally trained on a so-called ``two alternative forced choice'' (2AFC) dataset. Each data point of 2AFC contains three images: a reference, $p_0$, and $p_1$, where $p_0$ and $p_1$ are distorted in different ways based on the reference. A human evaluator then ranks $p_0$ and $p_1$ by their similarity to the reference. 
Here we propose the following modification to the dataset for training postprocess-specific metrics:
Similar to 2AFC, for each data point, we draw a reference image $x$ from the default generative model to be fingerprinted and define $p_0$ as the fingerprinted version of $x$. $p_1$ is then a postprocessed version of $x$ given a specific $T$ (or random combinations for \texttt{Combo}). To match with the setting of 2AFC, we sample 64 $\times$ 64 patches from $x$, $p_0$, and $p_1$ as training samples.
We then rank patches from $p_1$ as being more similar to those of $x$ than $p_0$. With this setting, the resulting LPIPS metric becomes more sensitive to fingerprints than mild postprocesses. 
The detailed training of the modified LPIPS follows the vgg-lin configuration in \cite{zhang2018unreasonable}.
It should be noted that, unlike previous SOTA where shallow fingerprint~\cite{kim2020decentralized} or encoder-decoder models~\cite{yu2020responsible} are retrained based on the known attacks, our fingerprinting mechanism, and therefore generalization performance, are agnostic to postprocesses.

\textbf{Accuracy-quality tradeoff.} 
We summarize fingerprint performance metrics on SG2 and FFHQ in Table~\ref{tab:robustness}. 
The attribution accuracies reported here are estimated using the \textbf{strongest} parameters of each attack.
For \texttt{Combo}, we use sequentially apply \texttt{Blurring}+\texttt{Noising}+\texttt{JPEG} as a deterministic worst-case attack. 
To estimate attribution accuracy, we solved the estimation problem in \eqref{eq:estimation} where postprocesses are applied.
\textbf{The proposed method:} We choose $V$ as a subset of 32 consecutive eigenvectors of $\Sigma_w$ starting from the 1th, 17th, and 33th eigenvectors, denoted respectively by PC[0:32], PC[16:48], and PC[32:64] in the table. fingerprinting strength $\sigma$ is set to 3. Attribution results from both a standard and a postprocess-specific LPIPS metric are reported in the UK (unknown) and KN (known) columns, respectively. Accuracies for our method are computed based on 100 random fingerprint samples from $2^{32}$, each with 100 random generations. \textbf{The baseline:} We compare with a shallow fingerprinting method from \cite{kim2020decentralized} (denoted by BL). When the postprocesses are known, BL performs postprocess-specific computation to derive shallow fingerprints that are optimally robust against the known postprocess. Results in UK and KN columns for BL are respectively without and with the postprocess-specific fingerprint computation. BL accuracies are computed based on 10 fingerprints, each with 100 random generations.

It is worth noting that the shallow fingerprinting method is not as scalable as ours \ck{(See Appendix:~\ref{appendix:runtime_comparison}}, and increasing the key capacity decreases the overall attribution accuracy (see \citep{kim2020decentralized}). Also, recall that the key length affects attribution accuracy (Proposition 1). Therefore, we conduct a fairer comparison to highlight the advantage of our method. Here we choose a subset of fingerprints $PC[32:40]$ (256 fingerprints) and report performance in Table~\ref{Tab:combination_sweep}, where accuracies are computed using the same settings as before. Visual comparisons between our method ($PC[32:40]$) and the baseline can be found in Fig.~\ref{fig:robustness}: To maintain attribution accuracy, high-strength shallow fingerprints, in the form of color patches, are needed around eyes and noses, and significantly lower the generation quality. In comparison, our method uses semantic changes that are robust to postprocesses. The choice of semantic dimensions, however, needs to be carefully chosen for the fingerprint to be perceptually subtle.

\textbf{Fingerprint secrecy.} 
The secrecy of the fingerprint is evaluated through the SSIM which is designed to measure the similarity between two images by taking into account three components: loss of correlation, luminance distortion, and contrast distortion~\cite{SSIM2004}. To facilitate a more equitable comparison between our methodology and the baseline, we select a subset of fingerprints presented in Table~\ref{Tab:combination_sweep} for evaluation. As shown in Table~\ref{Tab:combination_sweep}, the robustly fingerprinted model for $PC[32:40]$ exhibits superior fingerprint secrecy compared to the baseline, while simultaneously outperforming it in attribution accuracy. Besides these quantitative measures, our approach also demonstrates a qualitative advantage in terms of secrecy when compared to the baseline (see Fig.~\ref{fig:robustness}). This is largely attributed to the fact that subtle semantic variations across images are more challenging to visually detect and thus being removed compared to typical artifacts introduced by shallow fingerprinting.

\vspace{-0.05in}
\section{Conclusion}
\vspace{-0.05in}
This paper investigated latent fingerprint as a solution to enable the attribution of generative models. Our solution achieved a better tradeoff between attribution accuracy and generation quality than the previous SOTA that uses shallow fingerprints, and also has extremely low computational cost compared to SOTA methods that require encoder-decoder training with high data complexity, rendering our method more scalable to attributing large models with high-dimensional latent spaces.\\ 
\textbf{Limitations and future directions:} (1) There is currently a lack of certification on attribution accuracy due to the nonlinear nature of both the fingerprinting and the fingerprint estimation processes. Formally, by considering both the generation and estimation processes as discrete-time dynamics, such certification would require forward reachability analysis of fingerprinted contents and backward reachability analysis of the fingerprint, e.g., convex approximation of the support of $p_{x,i}$ and $\hat{\phi}$. It is worth investigating whether existing neural net certification methods can be applied. (2) Our method extracts fingerprints from the training data. Even with feature decomposition, the amount of features that can be used as fingerprints is limited. Thus the accuracy-quality tradeoff is governed by the data. It would be interesting to see if auxiliary datasets can help to learn novel and perceptually insignificant fingerprints that are robust against postprocesses, e.g., background patterns.

\vspace{-0.05in}
\section{Acknowledgment}
\vspace{-0.05in}
This work is partially supported by the National Science Foundation under Grant No. 2038666 and No. 2101052 and by an Amazon AWS Machine Learning Research Award (MLRA). Any opinions, findings, and conclusions expressed in this material are those of the author(s) and do not reflect the views of the funding entities.

\newpage
\bibliography{main}
\bibliographystyle{icml2023}

\newpage
\appendix
\onecolumn

\section{Proof of Propositions}
\subsection{Proposition 1} \label{appendix:proposition1}
Define $J_{w} = \nabla g(w)$, 
$H_{w} = J_{w}^TJ_{w}$, and $\bar{H}_{\phi} = \mathbb{E}_{\alpha \sim p_{\alpha}} [H_{U\alpha+\sigma V \phi}]$, where $p_{\alpha}$ is induced by $p_w$. Let $x_{\phi}(\alpha)$ be a content parameterized by $(\alpha, \phi)$. Denote by $\epsilon_{a} = \hat{a} - a$ the estimation error from the ground truth parameter $a$. Assume that the estimate $\hat{\alpha}(\alpha)$ is computed independent from $\phi$, and $\epsilon_{\alpha}$ is constant. Proposition 1 states:

\textbf{Proposition 1.}
\textit{$\exists c>0$ such that if $\sigma \leq c$ and $||\epsilon_{\alpha}||_2 \leq c$, the fingerprint estimation problem
\begin{equation} \label{eq:obj}
    \min_{\hat{\phi}} \quad \mathbb{E}_{\alpha \sim p_{\alpha}} \left[\|g(U\hat{\alpha}(\alpha)+\sigma V \hat{\phi}) - x_{\phi}(\alpha)\|_2^2\right]
\end{equation}
has an estimation error   
\begin{equation*} \label{eq:error}
    \epsilon_{\phi} = -(\sigma^2 V^T \bar{H}_{\phi} V)^{-1}V^T \bar{H}_{\phi} U \epsilon_{\alpha}.
\end{equation*}
}
\begin{proof}
Let $\hat{x} := g(U\hat{\alpha}_{\phi}(\alpha)+\sigma V \hat{\phi})$, we have
\begin{equation*}
    \hat{x} = g(U\hat{\alpha} + \sigma V \hat{\phi}) = g(U(\alpha + \epsilon_{\alpha}) + \sigma V (\phi + \epsilon_{\phi})).
\end{equation*}
With Taylor's expansion, we have
\begin{equation*} \label{eq:linear}
\begin{aligned}
    \hat{x} & = g(U\alpha + \sigma V \phi) + J_{w}(U\epsilon_{\alpha}+\sigma V \epsilon_{\phi}) + o(U\epsilon_{\alpha}+\sigma V \epsilon_{\phi}) \\
    & = x_{\phi}(\alpha) + J_{w}(U\epsilon_{\alpha}+\sigma V \epsilon_{\phi}) + o(U\epsilon_{\alpha}+\sigma V \epsilon_{\phi}).
\end{aligned}
\end{equation*}
Ignoring higher-order terms and , we then have
\begin{equation*} 
\begin{aligned}
    \|x_{\phi}(\alpha) - \hat{x}\|_2^2 & =  \|J_{w}(U\epsilon_{\alpha}+\sigma V \epsilon_{\phi}) + o(U\epsilon_{\alpha}+\sigma V \epsilon_{\phi})\|_2^2 \\
    & = \|J_{w}(U\epsilon_{\alpha}+\sigma V \epsilon_{\phi})\|_2^2 + o(U\epsilon_{\alpha}+\sigma V \epsilon_{\phi})^T J_{w}(U\epsilon_{\alpha}+\sigma V \epsilon_{\phi}).
\end{aligned}
\end{equation*}
For any $\tau > 0$, there exists $c$, such that if $\sigma \leq c$ and $||\epsilon_{\alpha}||_2 \leq c$, 
\begin{equation*}
\begin{aligned}
        \|x_{\phi}(\alpha) - \hat{x}\|_2^2 & \leq  \|J_{w}(U\epsilon_{\alpha}+\sigma V \epsilon_{\phi})\|_2^2 + \tau \\
        & = \sigma^2 \epsilon_{\phi}^T V^T H_{w} V \epsilon_{\phi} + 2 \epsilon_{\phi}^T V^T H_{w} U\epsilon_{\alpha} + \epsilon_{\alpha}^T U^T H_{w} U\epsilon_{\alpha} + \tau.
\end{aligned}
\end{equation*}
Removing terms independent from $\epsilon_{\phi}$ to reformulate \eqref{eq:obj} as
\begin{equation*} \label{eq:phi}
    \min_{\epsilon_{\phi}} 
    \quad 
    \sigma^2 \epsilon_{\phi}^T V^T \bar{H}_{\phi} V \epsilon_{\phi} + 2 \epsilon_{\phi}^T V^T \bar{H}_{\phi} U\epsilon_{\alpha},
\end{equation*}
the solution of which is
\begin{equation*} 
    \epsilon_{\phi} = -(\sigma^2 V^T \bar{H}_{\phi} V)^{-1}V^T \bar{H}_{\phi} U \epsilon_{\alpha}.
\end{equation*}

\end{proof}

\subsection{Proposition 2} \label{appendix:proposition2}
Consider two distributions: The first is $w_0 = \mu + U\alpha + V\beta$ where $\mu \in \mathbb{R}^{d_w}$, $\alpha \sim \mathcal{N}(0,diag(\lambda_U))$, and $\beta \sim \mathcal{N}(0,diag(\lambda_V))$. $diag(\lambda)$ is a diagonal matrix where diagonal elements follow $\lambda$. The second distribution is $w_1 = \mu + U\alpha + \sigma V\phi$ where $\sigma>0$ and $\phi \in \{0,1\}^{d_{\phi}}$. Let $g: \mathbb{R}^{d_w} \rightarrow \mathbb{R}^{d_x} \in C^1$. Let the mean and covariance matrix of $w_i$ be $\mu_i$ and $\Sigma_i$. Denote by $\bar{H}_U = \mathbb{E}_{\alpha}[J_{\mu + U\alpha}^T J_{\mu + U\alpha}]$ the mean Gram matrix, and let $\gamma_{U, max}$ be the largest eigenvalue of $\bar{H}_U$. Proposition 2 states:

\textbf{Proposition 2.} For any $\tau >0$ and $\eta \in (0,1)$, there exists $c(\tau, \eta)>0$ and $\nu>0$, such that if $\sigma \leq c(\tau, \eta)$ and $\lambda_{V,i} \leq c(\tau, \eta)$ for all $i=1,...,d_{\phi}$, $\|\mu_0 - \mu_1\|_2^2 \leq \sigma^2\gamma_{U, max} d_{\phi} + \tau$ and $|tr(\Sigma_0 - \Sigma_1)| \leq \lambda_{V, max} \gamma_{U, max} d_{\phi} + 2\nu \sigma \sqrt{d_{\phi}} + \tau$ with probability at least $1-\eta$.

\begin{proof}
We start with $\|\mu_0 - \mu_1\|_2^2$. From Taylor's expansion and using the independence between $\alpha$ and $\beta$, we have
\begin{equation} \label{eq:p2-1}
    \begin{aligned}
        \mu_0 := & \mathbb{E}_{\alpha, \beta}\left[ g(\mu + U\alpha + V\beta) \right]\\
        = & \mathbb{E}_{\alpha}\left[ g(\mu + U\alpha) \right] + \mathbb{E}_{\alpha, \beta}\left[ J_{\mu + U\alpha} V \beta + o(J_{\mu + U\alpha} V \beta) \right] + \\
        = & \mathbb{E}_{\alpha}\left[ g(\mu + U\alpha)\right] + \mathbb{E}_{\alpha, \beta}\left[ o(J_{\mu + U\alpha} V \beta)\right],\\
        \mu_1 := & \mathbb{E}_{\alpha}\left[ g(\mu + U\alpha + \sigma V\phi) \right]\\
        = & \mathbb{E}_{\alpha}\left[ g(\mu + U\alpha) + o(\sigma J_{\mu + U\alpha} V\phi) \right] + \sigma \mathbb{E}_{\alpha}\left[ J_{\mu + U\alpha} V \phi \right]\\
        = & \mu_0 + \sigma \mathbb{E}_{\alpha}\left[ J_{\mu + U\alpha} V \phi \right] + \mathbb{E}_{\alpha, \beta}\left[o(J_{\mu + U\alpha} V (\sigma \phi - \beta)) \right].
    \end{aligned}
\end{equation}
 
Let $v = V\phi$. With orthonormal $V$ and binary-coded $\phi$, we have 
\begin{equation} \label{eq:p2-2}
    \|v\|_2^2 = \phi^T V^TV \phi = \| \phi \|_2^2 \leq d_{\phi}.
\end{equation}
For the residual term $ \| \mathbb{E}_{\alpha, \beta}\left[o(J_{\mu + U\alpha} V (\sigma \phi - \beta)) \right] \|_2^2 $ and any $\tau >0$ and $\eta \in (0,1)$, there exists $c(\tau, \eta)>0$, such that if $\sigma \leq c(\tau, \eta)$ and $\lambda_{V,i} \leq c(\tau, \eta)$ for all $i=1,...,d_{\phi}$, we have
\begin{equation} \label{eq:p2-3}
    \Pr\left(\| \mathbb{E}_{\alpha, \beta}\left[o(J_{\mu + U\alpha} V (\sigma \phi - \beta)) \right] \|_2^2 \leq \tau\right) \geq 1-\eta.
\end{equation}
Lastly, we have
\begin{equation} \label{eq:p2-4}
\begin{aligned}
    \| \mathbb{E}_{\alpha}[J_{\mu + U\alpha}v] \|_2^2 & \leq \mathbb{E}_{\alpha}[v^T J_{\mu + U\alpha}^T J_{\mu + U\alpha} v] \\
    & = v^T \bar{H}_U v \leq \gamma_{U, max} \|v\|_2^2 \leq \gamma_{U, max} d_{\phi}.
\end{aligned}
\end{equation}
Then combining \eqref{eq:p2-1}, \eqref{eq:p2-2}, \eqref{eq:p2-3}, \eqref{eq:p2-4}, we have with probability at least $1-\eta$
\begin{equation}
    \|\mu_0 - \mu_1\|_2^2 \leq \sigma^2 \gamma_{U,max} d_{\phi} + \tau.
\end{equation}

For covariances, let $\Sigma_{U} = Cov(g(\mu + U\alpha))$. We have
\begin{equation}
\begin{aligned}
        \Sigma_0 := & \mathbb{E}_{\alpha, \beta}\left[ (g(\mu + U\alpha + V\beta) - \mu_0) (g(\mu + U\alpha + V\beta) - \mu_0) ^T\right] \\
        = & \Sigma_{U} + \mathbb{E}_{\alpha}\left[ J_{\mu + U\alpha} V diag(\lambda_V) V^T J_{\mu + U\alpha}^T \right] + \mathbb{E}_{\alpha, \beta}\left[ o(J_{\mu + U\alpha} V \beta)(g(\mu + U\alpha + V\beta) - \mu_0)^T\right]\\
        \Sigma_1 := & \mathbb{E}_{\alpha}\left[ (g(\mu + U\alpha + \sigma V\phi) - \mu_1) (g(\mu + U\alpha + \sigma V\phi) - \mu_1) ^T\right] \\
        = & \Sigma_{U} + \sigma^2 Cov(J_{\mu + U\alpha}V \phi^*) + 2\sigma Cov(g(\mu + U\alpha), J_{\mu + U\alpha}V \phi + o(J_{\mu + U\alpha}V \phi)).
\end{aligned}
\end{equation}

For $tr(\Sigma_0)$, using the same treatment for the residual, we have for any $\tau>0$ and $\eta \in (0,1)$, there exists $c(\tau, \eta)>0$, such that if $\lambda_{V,i} \leq c(\tau, \eta)$ for all $i=1,...,d_{\phi}$, the following upper bound applies with at least probability $1-\eta$:
\begin{equation}
\begin{aligned}
    tr(\Sigma_0) & \leq tr(\Sigma_{U}) + tr(\mathbb{E}_{\alpha}\left[ J_{\mu + U\alpha} V diag(\lambda_V) V^T J_{\mu + U\alpha}^T \right]) + \tau \\
    & \leq tr(\Sigma_{U}) + \lambda_{V, max} tr(\mathbb{E}_{\alpha} \left[ J_{\mu + U\alpha} VV^T J_{\mu + U\alpha}^T \right]) + \tau  \\
    & = tr(\Sigma_{U}) + \lambda_{V, max} tr(\mathbb{E}_{\alpha} \left[V^T J_{\mu + U\alpha}^T J_{\mu + U\alpha} V \right]) + \tau \\
    & \leq tr(\Sigma_{U}) + \lambda_{V, max} \gamma_{U, max} tr(V^TV) + \tau \\
    & \leq tr(\Sigma_{U}) + \lambda_{V, max} \gamma_{U, max} d_{\phi} + \tau.
\end{aligned}
\end{equation}
For the lower bound, we have $tr(\Sigma_0) \geq tr(\Sigma_{U})$.

For $tr(\Sigma_1)$, we first denote by $J_i^T$ the $i$th row of $J_{\mu+U\alpha}$, $\Sigma_{J_i}$ its covariance matrix, and $\sigma_i^2$ the maximum eigenvalue of $\Sigma_{J_i}$. Then with binary-coded $\phi$, we have
\begin{equation} \label{eq:p2-1}
    Var(J_i^T V \phi) = \phi^T V^T Cov(J_i) V \phi \leq \sigma_i^2 d_{\phi}.
\end{equation}
Then let $g_i$ (resp. $v_i$) be the $i$th element of $g(\mu+U\alpha)$ (resp. $J_{\mu + U\alpha}V\phi$), and $\sigma_{U,i}^2$ be the $i$th diagonal element of $\Sigma_U$. Using \eqref{eq:p2-1}, we have the following bound on the trace of the covariance between $g(\mu+U\alpha)$ and $J_{\mu + U\alpha}V\phi$:
\begin{equation}
    \left|tr(Cov(g(\mu+U\alpha), J_{\mu + U\alpha}V\phi))\right| = \left|\sum_{i=1}^{d_x} Cov(g_i, v_i)\right| \leq \sum_{i=1}^{d_x} \sigma_{U,i}\sigma_i \sqrt{d_{\phi}}.
\end{equation}

Lastly, by ignoring $\sigma^2$ terms and borrowing the same $\tau$, $\eta$, and $c(\tau,\eta)$, we have with probability at least $1-\eta$:
\begin{equation}
\begin{aligned}
    tr(\Sigma_1) & \leq tr(\Sigma_{U}) + 2\sigma tr(Cov(g(\mu+U\alpha), J_{\mu + U\alpha}V\phi)) + \tau \\
    & \leq tr(\Sigma_{U}) + 2\sigma\sum_{i=1}^{d_x} \sigma_{U,i}\sigma_i \sqrt{d_{\phi}} + \tau,
\end{aligned}
\end{equation}
and 
\begin{equation}
    tr(\Sigma_1) \geq tr(\Sigma_{U}) - 2\sigma \sum_{i=1}^{d_x} \sigma_{U,i}\sigma_i \sqrt{d_{\phi}}.
\end{equation}
Therefore, with probability at least $1-\eta$
\begin{equation}
    tr(\Sigma_0) - tr(\Sigma_1) \leq \lambda_{V, max} \gamma_{U, max} d_{\phi} + 2\sigma \sum_{i=1}^{d_x} \sigma_{U,i}\sigma_i \sqrt{d_{\phi}} + \tau, 
\end{equation}
and
\begin{equation}
    tr(\Sigma_0) - tr(\Sigma_1) \geq -2\sigma \sum_{i=1}^{d_x} \sigma_{U,i}\sigma_i \sqrt{d_{\phi}} - \tau. 
\end{equation}
\end{proof}



\section{Convergence on $\alpha$}
\label{app:error_alpha}
\begin{figure}
\centering
\includegraphics[width=\linewidth]{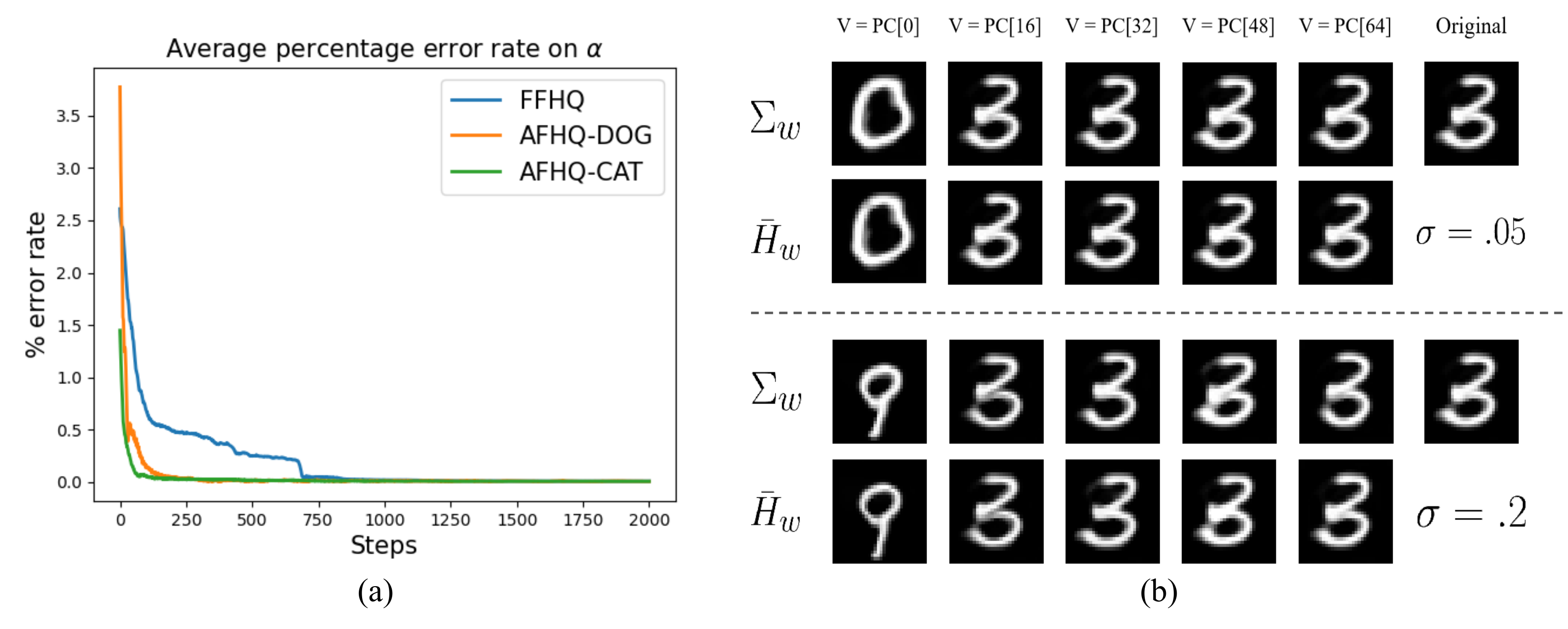}
\caption{(a) Average percentage error rate on $\alpha$ (b) Comparison between fingerprints guided by $\Sigma_w$ and $\bar{H}_{w}$. The editing strength for top two rows and bottom two rows are 0.05 and 0.2 respectively.}
\label{fig:alpha_error_rate_and_gram_sig}
\end{figure}

In the proofs, we assume that $\|\epsilon_{\alpha}\|_2$ is small and constant. Here we show empirical estimation results on SG2 and on FFHQ, AFHQ-DOG, AFHQ-CAT datasets. The results in Fig.~\ref{fig:alpha_error_rate_and_gram_sig}(a) are averaged over 100 random $\alpha$ and 100 random $\phi$, and uses parallel search on $\alpha$ during the estimation.

\section{Qualitative similarity between $\bar{H}_w$ and $\Sigma_w$}
\label{app:similar}

Since computing $\bar{H}_w$ for large models is intractable, here we train a SG2 on MNIST to estimate $\bar{H}_w$. Fig.~\ref{fig:alpha_error_rate_and_gram_sig}(b) summarizes perturbed images from a randomly chosen reference along principal components of $\bar{H}_w$ and $\Sigma_w$. Note that both have quickly diminishing eigenvalues. Therefore most components other than the few major ones lead to imperceptible changes in the image space.

\section{Computational Complexity and Efficiency Comparison of Proposed and Baseline Methods} \label{appendix:runtime_comparison}
To comprehensively evaluate the computational costs of the proposed and baseline methods, we analyzed them from two perspectives: fingerprint generation and attribution. Our proposed method demonstrates a significant increase in efficiency compared to existing methods with regards to fingerprint generation. However, current methods, such as those proposed in~\cite{kim2020decentralized}, often necessitate a considerable amount of time for fine-tuning the model before generating a fingerprinted image. For instance, the baseline method requires approximately one hour to fine-tune the model for each individual user. For a key capacity of $2^{32}$ users, it would take an estimated $2^{32}$ hours to train on an NVIDIA V100 GPU. In contrast, our proposed method does not require the model to be fine-tuned, and only requires Principal Component Analysis (PCA) to be performed on the latent space of each pre-trained model to identify the editing direction. This process takes approximately three seconds for a key capacity of $2^{32}$.

Regarding attribution time, the baseline approach performs attribution through a pre-trained network, resulting in low computation cost during inference. On the other hand, the proposed method uses optimization techniques to perform attribution, resulting in a longer computation time. Specifically, the average attribution time for the baseline method is approximately two seconds, while the proposed method takes approximately 126 seconds on average for 1k optimization trials (in parallel) using an NVIDIA V100 GPU.

It is important to acknowledge that there exists a trade-off between computational costs during generation and attribution. The choice of which aspect is more critical may depend on the specific application requirements.

\section{Ablation Study }\label{appendix:sec:stegastamp_included_baseline}
\begin{figure}[p]
    \centering
    \includegraphics[width=12cm]{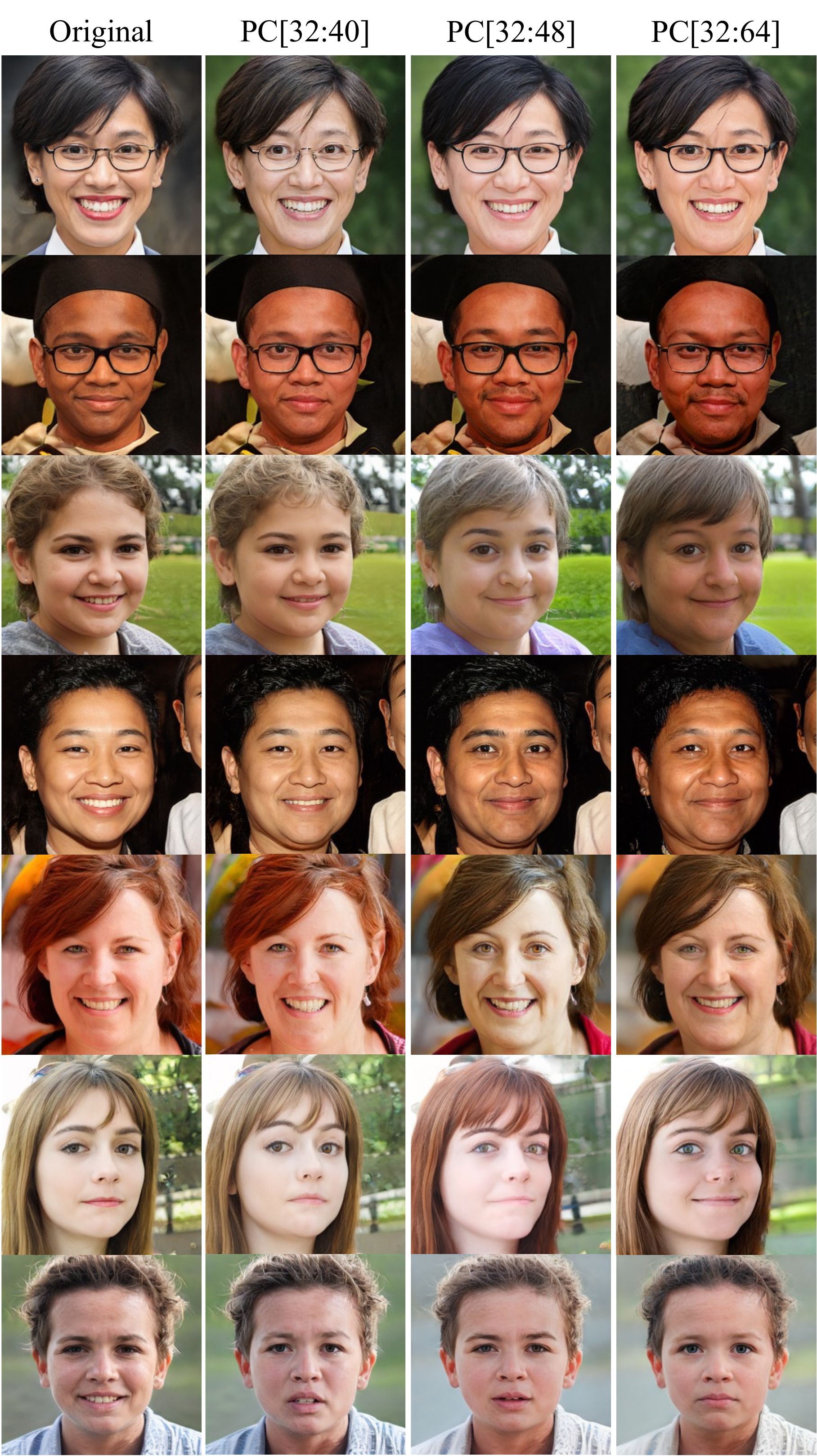}
    \caption{Qualitative Comparison of Fingerprinted Samples. The first column shows original images $g_{0}(w)$. 
    Latent fingerprinted images are in the second to the last column.
    }%
    \label{fig:new_baseline}%
\end{figure}

In this section, we estimated attribution accuracy based on various attack parameters with multiple editing directions (see Tab.\ref{tab:blur},\ref{tab:noise},\ref{tab:jpeg},\ref{tab:combination}).
The image quality evaluation is available in Tab.\ref{tab:quality_comparison} and more visualizations can be found in Fig.\ref{fig:new_baseline}.

\begin{table} [h]
  \caption{
  Attributability Table of \texttt{Blurring} attack. $\sigma$ refers standard deviation of Gaussian Blur filter size 25. When attributability is measured with (without) knowledge of attack, we put results under KN (UK).
  }
  \label{tab:blur}
  \centering
  \renewcommand{\tabcolsep}{5pt}
  \begin{tabular}{llllllllll}
    \toprule
    Metric & Model &\multicolumn{2}{c}{\texttt{$\sigma$=0.5}} &  \multicolumn{2}{c}{\texttt{\texttt{$\sigma$=1.0}}} & \multicolumn{2}{c}{\texttt{\texttt{$\sigma$=1.5}}} & \multicolumn{2}{c}{\texttt{\texttt{$\sigma$=2.0}}}\\
    -      & -  & UK & KN & UK & KN & UK & KN & UK & KN \\ 
    \midrule
    \multirow{5}{*}{Att. $\uparrow$} 
    & BL  & 0.88 & 0.89  &0.87 & 0.89 & 0.87 & 0.88 &0.85 &0.88 \\
    & PC[32:40]  & 0.99 & 0.99  &0.95 &0.99  & 0.90 & 0.99 &0.53 &0.92 \\
    & PC[32:48] & 0.99 &0.99 &0.97 & 0.99 & 0.72 & 0.92 & 0.38 & 0.88\\
    & PC[32:64]  & 0.99 & 0.99  &0.73 & 0.94 & 0.51 & 0.90 &0.32 &0.83 \\
    
    \bottomrule
    
  \end{tabular}
\end{table}


\begin{table} [h]
  \caption{
  Attributability Table of \texttt{Noise} attack. $\sigma$ refers standard deviation of Gaussian normal distribution. When attributability is measured with (without) knowledge of attack, we put results under KN (UK). 
  }
  \label{tab:noise}
  \centering
  \renewcommand{\tabcolsep}{5pt}
  \begin{tabular}{llllllllll}
    \toprule
    Metric & Model &\multicolumn{2}{c}{\texttt{$\sigma$=0.025}} &  \multicolumn{2}{c}{\texttt{\texttt{$\sigma$=0.05}}} & \multicolumn{2}{c}{\texttt{\texttt{$\sigma$=0.075}}} & \multicolumn{2}{c}{\texttt{\texttt{$\sigma$=0.1}}}\\
    -      & -  & UK & KN & UK & KN & UK & KN & UK & KN \\ 
    \midrule
    \multirow{3}{*}{Att. $\uparrow$} 
    & BL  & 0.87 & 0.88  &0.86 & 0.88 & 0.86 & 0.87 &0.85 &0.87 \\
    & PC[32:40]  & 0.99 & 0.99  &0.99 & 0.99 & 0.99 & 0.99 &0.99 &0.99 \\
    & PC[32:48] & 0.99 &0.99 &0.97 & 0.99 & 0.94 & 0.99 & 0.95 & 0.99\\
    & PC[32:64]  & 0.98 & 0.99  &0.95 & 0.99 & 0.92 & 0.98 &0.93 &0.98 \\

     \bottomrule
    
  \end{tabular}
\end{table}


\begin{table} [h]
  \caption{
  Attributability Table of JPEG attack. Q refers quality metric of JEPG compression. When attributability is measured with (without) knowledge of attack, we put results under KN (UK).
  }
  \label{tab:jpeg}
  \centering
  \renewcommand{\tabcolsep}{5pt}
  \begin{tabular}{llllllllll}
    \toprule
    Metric & Model &\multicolumn{2}{c}{\texttt{Q=80}} &  \multicolumn{2}{c}{\texttt{\texttt{Q=70}}} & \multicolumn{2}{c}{\texttt{\texttt{Q=60}}} & \multicolumn{2}{c}{\texttt{\texttt{Q=50}}}\\
    -      & -  & UK & KN & UK & KN & UK & KN & UK & KN \\ 
    \midrule
    \multirow{3}{*}{Att. $\uparrow$} 
    & BL  & 0.88 & 0.89  &0.87 & 0.89 & 0.87 & 0.89 &0.87 &0.89 \\
    & PC[32:40]  & 0.99 & 0.99  &0.99 & 0.99 & 0.99 & 0.99 &0.99 &0.99 \\
    & PC[32:48] & 0.99 &0.99 &0.99 & 0.99 & 0.99 & 0.99 & 0.99 & 0.99\\
    & PC[32:64]  & 0.99 & 0.99  &0.99 & 0.99 & 0.99 & 0.99 &0.98 &0.99 \\

     \bottomrule
    
  \end{tabular}
\end{table}


\begin{table} [h]
  \caption{
  Attributability Table of combination attack. From T1 to T4, the attack parameters are composed of the weakest to the strongest attack parameters of each attack (e.g., T4 is [$\sigma_{blur}=2.0$, $\sigma_{noise}=0.2$, Q\textsubscript{JPEG}=50]).
  When attributability is measured with (without) knowledge of attack, we put results under KN (UK).
  }
  \label{tab:combination}
  \centering
  \renewcommand{\tabcolsep}{5pt}
  \begin{tabular}{llllllllll}
    \toprule
    Metric & Model &\multicolumn{2}{c}{\texttt{T1}} &  \multicolumn{2}{c}{\texttt{\texttt{T2}}} & \multicolumn{2}{c}{\texttt{\texttt{T3}}} & \multicolumn{2}{c}{\texttt{\texttt{T4}}}\\
    -      & -  & UK & KN & UK & KN & UK & KN & UK & KN \\ 
    \midrule
    \multirow{3}{*}{Att. $\uparrow$} 
    & BL  & 0.86 & 0.88  &0.86 & 0.87 & 0.85 & 0.86 &0.83 &0.88 \\
    & PC[32:40]  & 0.99 & 0.99  &0.94 & 0.99 & 0.81 & 0.95 &0.65 &0.89 \\
    & PC[32:48] & 0.99 &0.99 &0.74 & 0.92 & 0.52 & 0.88 & 0.45 & 0.85\\
    & PC[32:64]  & 0.99 & 0.99  &0.63 & 0.90 & 0.41 & 0.82 &0.26 &0.79 \\

     \bottomrule
    
  \end{tabular}
\end{table}

\begin{table} [h]
  \caption{
  Quality Comparison Table. Standard deviation are in parentheses. The baseline score is in the parentheses. 
  }
  \label{tab:quality_comparison}
  \centering
  \renewcommand{\tabcolsep}{5pt}
  \begin{tabular}{llll}
    \toprule
    Model &FID (7.24)$\downarrow$
    & IS (4.95)$\uparrow$
    & SSIM $\uparrow$
    \\
    \midrule
    BL\textsubscript{Blur}
     & 99.05
     & 2.86 (0.35)
     & 0.67(0.08)
    \\
    BL\textsubscript{Noise}
     & 93.04
     & 3.02 (0.27)
     & 0.68(0.07)
    \\
    BL\textsubscript{JPEG}
     & 97.70
     & 2.91 (0.26)
     & 0.67(0.07)
    \\
    Bl\textsubscript{Combo}
     & 100.15
     & 2.90 (0.23)
     & 0.66(0.06)
    \\
    PC[32:40]
     & \textbf{12.35}
     & 4.75 (0.05)
     &\textbf{0.73(0.06)}
    \\
    PC[32:48]
     & 13.25
     & \textbf{4.86 (0.08)}
     & 0.65(0.06)
    \\
    PC[32:64]
     & 27.50
     & 4.50 (0.07)
     & 0.57(0.07)
    \\
    
     \bottomrule
    
  \end{tabular}
\end{table}

\end{document}